\documentclass{article}
\usepackage[square,numbers]{natbib}
\usepackage{authblk}
\usepackage{charter}
\usepackage{fullpage}
\usepackage{bbm}
\usepackage[toc,page]{appendix}
\usepackage[most]{tcolorbox}
\usepackage[ruled,vlined]{algorithm2e}
\usepackage{titletoc}
\usepackage{minitoc}

\usepackage{hyperref}       
\usepackage{url}            
\usepackage{booktabs}       
\usepackage{amsfonts}       
\usepackage{nicefrac}       
\usepackage{microtype}      
\usepackage[table]{xcolor}         
\usepackage{amsmath,amssymb,amsthm,amsfonts}
\usepackage{latexsym,bbm,graphicx,float,mathtools}
\usepackage{algorithmic}
\usepackage[export]{adjustbox}
\usepackage{capt-of}
\usepackage{booktabs}
\usepackage{parskip}

\usepackage[utf8]{inputenc}

\usepackage[noabbrev]{cleveref}
\usepackage{caption}
\usepackage{subcaption}
\usepackage{transparent}
\usepackage[inline]{enumitem}
\usepackage{multirow}
\usepackage{multicol}
\usepackage{xspace}

\usepackage{abstract}

\usepackage[subtle, mathdisplays=tight, charwidths=normal, leading=normal]{savetrees}

\newcommand{\name}{\textsc{EditBridge: }\xspace}

\title{\textbf{\name Towards Faithful and Efficient \\ Ultra-High-Resolution Image Editing}}

\author{
    Jiayi Song\textsuperscript{1,2},
    Shijie Huang\textsuperscript{2},
    Fangtai Wu\textsuperscript{2},
    Yubo Huang\textsuperscript{2},
    Zhenxiong Tan\textsuperscript{3}
\par \vspace{-0.2ex} 
    Songhua Liu\textsuperscript{1}\thanks{Corresponding Author.} ,
    Jiaming Liu\textsuperscript{2}\thanks{Project Lead.} ,
    Ruihua Huang\textsuperscript{2}
   \\ \vspace{-0.3ex}
\textsuperscript{1}School of Artificial Intelligence, Shanghai Jiao Tong University \\
  \textsuperscript{2} Qwen Business Unit of Alibaba \\
  \textsuperscript{3}National University of Singapore \\
  \texttt{liusonghua@sjtu.edu.cn}
  }

\begin{document}

\maketitle

\begin{abstract}
High-resolution image editing is increasingly demanded in professional 
workflows, yet existing diffusion-based models remain constrained to 
resolutions below 1K due to quadratic attention complexity and prohibitive 
memory requirements. A prevalent workaround employs a two-stage pipeline: editing at low resolution followed by independent super-resolution. However,  
this approach suffers from two critical issues: \textit{information divergence}, 
where hallucinated details contradict the original high-resolution (HR) source, 
and \textit{texture degradation}, manifesting as over-smoothed or over-sharpened 
artifacts. We propose \textbf{EditBridge}, a diffusion bridge framework for 
efficient ultra high-resolution editing. Unlike conventional diffusion that 
regenerates from noise, we formulate refinement as structured data-to-data 
translation from the low-resolution (LR) edited result to its HR counterpart, 
explicitly conditioned on the original HR source to preserve authentic details. 
To efficiently incorporate HR source guidance, we introduce a prior-guided 
block-wise sparse attention mechanism that exploits semantic correspondence 
from first-stage editing to constrain cross-image interactions to spatially 
aligned regions, significantly reducing computational overhead. Extensive 
experiments demonstrate that EditBridge achieves high-fidelity editing with 
superior perceptual quality at resolutions up to 4K, delivering 3.6--8.4$\times$ speedup at 2K and enabling practical 4K editing in 61 seconds. Project page: \url{https://editbridge.github.io/}.
\end{abstract}

\section{Introduction}

High-resolution visual content generation has become a central topic in modern generative modeling, spanning both image~\cite{bu2025hiflow,ye2025ultraflux,qiu2025freescale} and video~\cite{skorokhodov2024hierarchical,blattmann2023align} synthesis. However, due to the quadratic computational complexity of attention mechanisms and substantial memory requirements~\cite{tokemerge,attention_is_all_your_need}, most existing image editing models~\cite{kontext,bagel,wu2025qwenimagetechnicalreport,superintelligenceteam2026fireredimageedit10technicalreport}
remain constrained to resolutions no higher than 1K ($1024 \times 1024$) during inference. To obtain higher-resolution outputs, a common paradigm relies on a two-stage pipeline: performing the edit at a lower resolution and subsequently employing an independent super-resolution(SR) model for upscaling~\cite{cheng2025effective}.

\begin{figure}[t]
    \centering
    \includegraphics[width=1\linewidth]{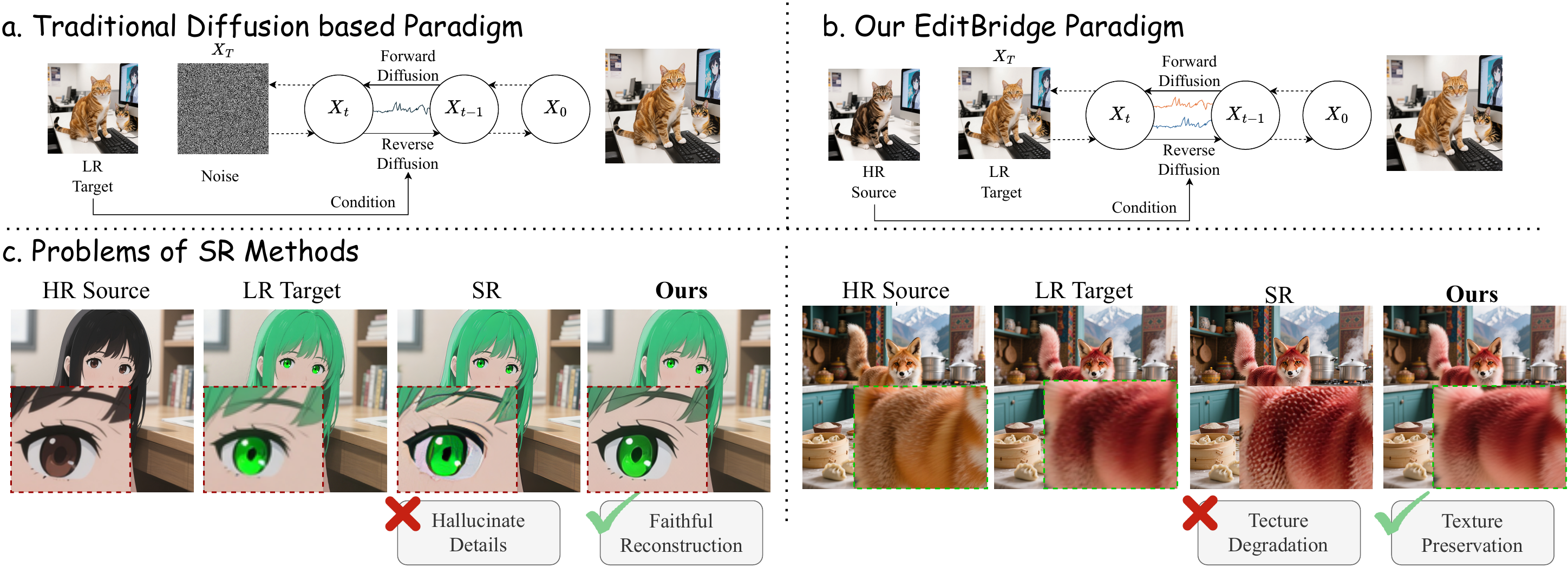}
    \caption{Motivation of the proposed method. (a) and (b) compare the standard high-resolution image editing paradigm with our approach. (c) analyzes the inherent limitations of existing methods.}
    \vspace{-8mm}
    \label{fig:motivation}
\end{figure}

While super-resolution increases the output resolution, it tends to hallucinate 
high-frequency textures due to the absence of high-resolution(HR) source guidance. This introduces 
two primary challenges: (1) \textit{Information Divergence}: the hallucinated details inevitably diverge 
from the original HR source content, contradicting the fundamental goal of faithful 
editing—as shown in Fig.~\ref{fig:motivation}(c) left,
where the facial details produced by SR are entirely inconsistent with the HR source; 
and (2) \textit{Texture Degradation}: the generation of over-smoothed or over-sharpened 
artifacts that degrade visual fidelity, as illustrated in Fig.~\ref{fig:motivation}(c) 
right.

A straightforward alternative would be to train an image editing model natively at high resolutions. However, such an approach demands exorbitant computational resources and large-scale, high-resolution editing datasets, making it prohibitively expensive in practice. Moreover, even with a trained model, high-resolution inference remains computationally intensive due to the quadratic complexity of attention 
mechanisms, resulting in severe efficiency bottlenecks.  This naturally raises a key question: \textit{Can we achieve high-resolution image editing in a more faithful and efficient manner?}

Motivated by this question, we propose \textbf{EditBridge}, a prior-guided 
diffusion bridge framework for ultra-high-resolution image editing. Since 
the core limitation of SR methods lies in the absence of HR source guidance, 
our key idea is to explicitly leverage the original high-resolution source 
image as conditional guidance during refinement. EditBridge addresses this 
through two core designs: (1) a diffusion bridge formulation that models 
the structured refinement from LR edited images to HR outputs, and (2) a 
prior-guided sparse attention mechanism that efficiently incorporates HR 
source information without prohibitive computational overhead.

Specifically, we formulate high-resolution editing as a refinement process 
that transforms the low-resolution (LR) edited image into its high-resolution 
(HR) counterpart. Unlike generic upsampling that operates solely on the LR 
input, our goal is to synthesize fine-grained textures guided by the original 
HR source while preserving consistency with the edited semantic content.

A key observation is that this task differs fundamentally from standard 
conditional generation: we have a concrete LR edited image as the starting 
point, not random noise. This motivates us to adopt diffusion 
bridges~\cite{li2023bbdm,zhoudenoising}, which model stochastic transitions 
between two data distributions rather than from noise to data. Compared to 
conventional diffusion models that must ``re-generate'' the entire image 
from scratch, diffusion bridges preserve the structural information from 
the LR input throughout the refinement trajectory, enabling more faithful 
and efficient high-resolution synthesis.

While the diffusion bridge provides a principled formulation, a key challenge 
remains: how to efficiently incorporate the HR source image as guidance. A 
naive approach would concatenate contextual tokens directly within the 
Diffusion Transformer (DiT)~\cite{peebles2023scalable}, but this incurs 
prohibitive computational overhead due to token expansion and the quadratic 
complexity of global attention. We observe that high-resolution editing does not necessitate exhaustive all-to-all attention between target  queries and source keys. Instead, interactions should be governed by semantic relevance: for unedited regions, the model primarily requires high-fidelity preservation, which can be achieved by attending strictly to semantically corresponding anchors in the HR source image; for edited regions, the model must synthesize new high-frequency details by assimilating local contextual cues.

Building upon this insight, we propose a prior-guided block-wise sparse attention mechanism that explicitly routes information based on spatial semantic alignment. During the LR source-to-target editing stage, our approach extracts semantic correspondence priors from the attention maps. These priors serve as a spatial roadmap, identifying the precise regions relevant to the modification and locating their corresponding anchors in the HR source. Guided by these cues, the sparse attention mechanism constrains the HR receptive field to aligned areas, effectively pruning redundant interactions with irrelevant tokens. Consequently, EditBridge achieves a superior balance between computational efficiency and high-fidelity texture synthesis.

In summary, the main contributions of this paper are:
\begin{itemize}
    \item We present a DiT-based framework for high-resolution image editing that 
achieves both computational efficiency and faithful detail preservation.
    \item We propose a diffusion-bridge-based refinement stage equipped with a prior-guided block-wise sparse attention mechanism, facilitating semantically selective cross-image interaction.
    \item Our method achieves state-of-the-art high-resolution editing performance while drastically reducing computational costs, yielding up to a $2.2\times$ speedup compared to conventional diffusion-based approaches.
\end{itemize}
\section{Related Work}
\textbf{Instruction Based Image Editing.} 

The rapid advancement of diffusion models has significantly advanced instruction-based image editing. Early works such as InstructPix2Pix~\cite{brooks2023instructpix2pix} and SuTI~\cite{chen2023subject} introduced the paradigm of editing images directly from instructions rather than descriptive captions. MagicBrush~\cite{zhang2023magicbrush} further improved editing capabilities by introducing high-quality, diverse training datasets. Beyond dataset-driven improvements, subsequent approaches~\cite{huang2024smartedit,fu2023guiding,li2023instructany2pix} integrated Multimodal Large Language Models (MLLMs)~\cite{mllm_survey,qwenvl3,qwenvl25} with diffusion models~\cite{bagel,wu2025qwenimagetechnicalreport} to enhance instruction understanding and controllability. With the introduction of Diffusion Transformers (DiT)~\cite{peebles2023scalable}, research efforts have shifted toward developing unified DiT-based foundation models for conditional image generation ~\cite{zhang2025easycontroladdingefficientflexible,tan2025ominicontrolminimaluniversalcontrol} and editing~\cite{kontext,bagel,wu2025qwenimagetechnicalreport,superintelligenceteam2026fireredimageedit10technicalreport}, injecting conditional signals directly into attention layers for fine-grained guidance. Despite this flexibility, these approaches incur substantial computational overhead due to dense global attention operations, a bottleneck that becomes prohibitive at high resolutions.

\textbf{High-Resolution Visual Generation.}
Recent works have explored high-resolution visual generation~\cite{bu2025hiflow,ye2025ultraflux,du2024max}. Early approaches~\cite{blattmann2023align,blattmann2023stable} primarily rely on U-Net~\cite{ronneberger2015u} architectures. With the rise of foundation diffusion models, DiT~\cite{peebles2023scalable} has become the dominant backbone, as its attention mechanism excels at modeling complex token-wise dependencies. To improve inference efficiency at high resolutions, several acceleration methods have emerged. For instance, I-Max~\cite{du2024max} projects high-resolution flows into a lower-resolution latent space to reduce computational costs. HiFlow~\cite{bu2025hiflow} introduces a model-agnostic acceleration framework for scalable generation, while CLEAR~\cite{liuclear} proposes a linearization strategy tailored  to mitigate the quadratic complexity of global attention.

However, these methods are primarily designed for pure image generation tasks. In contrast, image editing strictly requires fine-grained consistency between the source and target images, rendering generation-oriented acceleration strategies sub-optimal for editing scenarios. To bridge this gap, works like MobilePicasso~\cite{kwon2025efficienthighresolutionimageediting} introduce hybrid pipelines with hallucination-aware training and adaptive tiling. Similarly, ScaleEdit~\cite{lee2025lowresolutioneditingneedhighresolution} proposes a training-free transfer function for high-resolution editing. However, these methods are built upon U-Net architectures and cannot be 
directly applied to the increasingly prevalent DiT-based editing frameworks. In this work, we address this gap by proposing EditBridge, a diffusion 
bridge framework specifically designed for efficient and faithful 
high-resolution editing within the DiT paradigm.

\begin{figure}[t]
    \centering
    \includegraphics[width=0.95\linewidth]{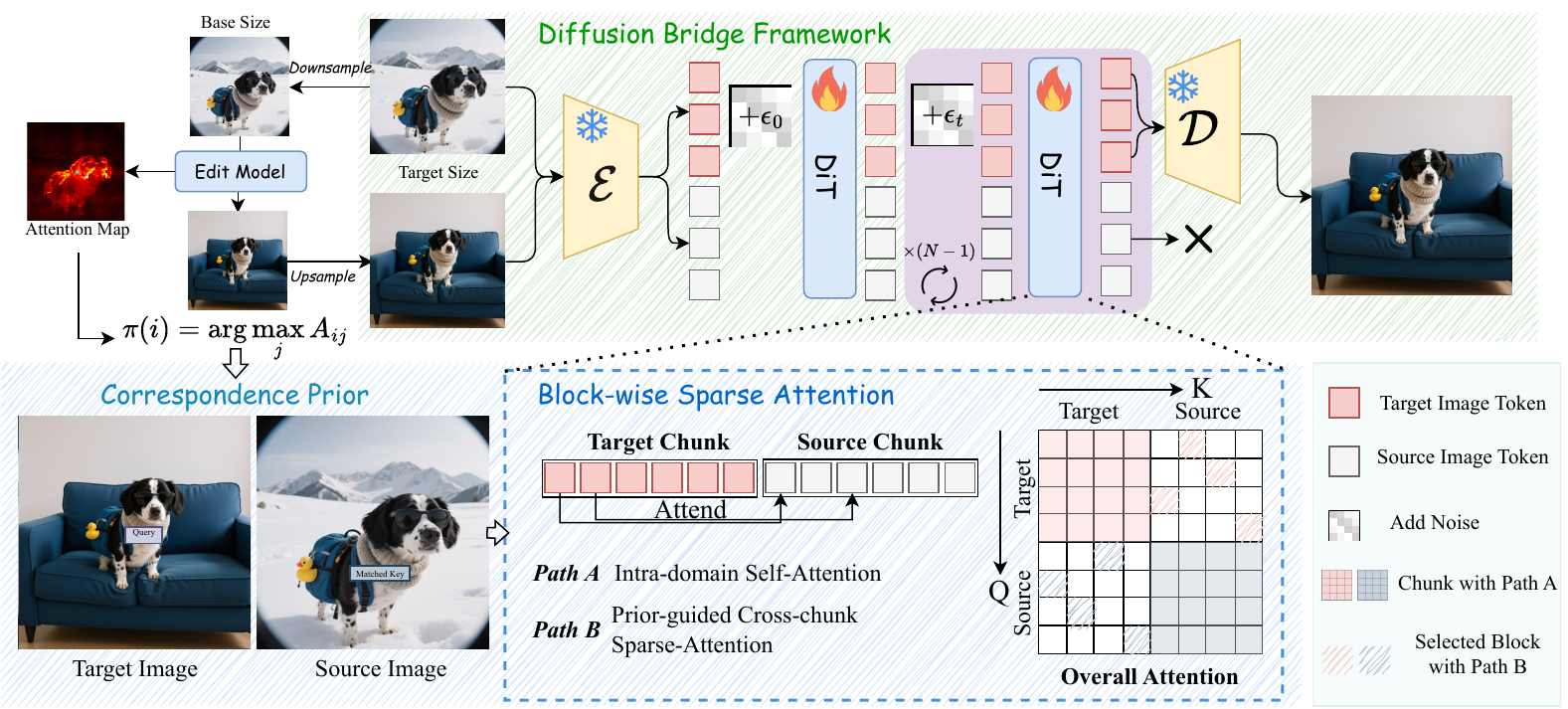}
    \caption{\textbf{Overview of our proposed EditBridge.} The upper section illustrates the inference process of the diffusion bridge, which transports the low-resolution edit to its high-resolution counterpart. The lower section details the construction of the correspondence prior and the proposed Prior-Guided Sparse Attention (PG-BSA) mechanism.}
    \label{fig:method}
    
\end{figure}
\section{Preliminary: Diffusion Bridge}
\label{sec:preliminary}
\textbf{Diffusion Bridge.} Standard diffusion models learn to transport samples from an uninformative 
prior (e.g., Gaussian noise) to the data distribution. In contrast, 
\textit{diffusion bridges} (also known as Brownian bridges) model stochastic 
paths between two structured data domains, making them naturally suited for 
data-to-data translation tasks. Formally, given paired endpoints $(x_0, x_1)$ from source and target domains, 
the Brownian bridge defines a stochastic interpolation path. The intermediate 
state $X_t$ at time $t \in [0,1]$ follows:
\begin{equation}
    X_t \mid (x_0, x_1) \sim \mathcal{N}\left( (1-t)x_0 + t x_1,\; t(1-t)I \right).
\end{equation}
Intuitively, $X_t$ is a noisy interpolation between the two endpoints, with 
the noise level peaking at $t=0.5$ and vanishing at the boundaries. The corresponding instantaneous velocity that drives this transition is:
\begin{equation}
    u_t(X_t \mid x_0, x_1) = \frac{x_1 - X_t}{1 - t}.
\end{equation}

In practice, we parameterize a neural network $v_\theta$ to approximate this 
velocity field via the matching objective:
\begin{equation}
    \mathcal{L}(\theta) = \mathbb{E}_{x_0, x_1, t, X_t} \left\| v_\theta(X_t, t) - u_t(X_t \mid x_0, x_1) \right\|^2.
    \label{eq:loss}
\end{equation}

Compared to standard diffusion that reconstructs signals from 
noise, the bridge formulation preserves structural information from $x_0$ 
throughout the trajectory, enabling efficient and faithful transformations. 

\textbf{Problem Formulation.}
High-resolution (HR) image editing aims to map an HR source $x_s^{HR}$ and instruction $c$ to an edited HR output $x_t^{HR}$. To bypass the computational burden of direct HR optimization, we adopt a coarse-to-fine paradigm:
\begin{enumerate}
    \item \textbf{LR Editing:} $x_t^{LR} = \mathcal{G}(x_s^{LR}, c)$, where $\mathcal{G}$ is a pre-trained low-resolution (LR) model and $x_s^{LR}$ is the downsampled source.
    \item \textbf{HR Refinement:} $x_t^{HR} = \mathcal{H}(\tilde{x}_t^{HR}, x_s^{HR})$, where $\tilde{x}_t^{HR}$ is the upsampled $x_t^{LR}$.
\end{enumerate}
Our objective is to learn the refinement mapping $\mathcal{H}$. Notably, 
this HR refinement task naturally fits the diffusion bridge formulation: 
the upsampled LR edit $\tilde{x}_t^{HR}$ serves as the source endpoint $x_0$, 
and the target HR output $x_t^{HR}$ serves as the target endpoint $x_1$.
\section{Method}

Based on the formulation, we present two key components of our 
framework: Bridge-based HR Refinement (Sec.~\ref{sec:bridge_refine}) and 
Prior-Guided Sparse Attention (Sec.~\ref{sec:sparse_attention}). An overview 
is shown in Fig.~\ref{fig:method}.

\subsection{Bridge-based High-Resolution Refinement}
\label{sec:bridge_refine}
As illustrated in Fig.~\ref{fig:motivation}b, based on the diffusion bridge formulation introduced in Sec.~\ref{sec:preliminary}, we instantiate a conditional bridge model tailored for high-resolution refinement. In our framework, the source endpoint $x_0$ is defined as upsampled low-resolution edited result $\tilde{x}_t^{HR}$. The target endpoint $x_1$ corresponds to the desired high-resolution edited image $x_t^{HR}$. The objective of the second stage is to model a probability path that characterizes the transition from $\tilde{x}_t^{HR}$ to $x_t^{HR}$.

Unlike unconditional generative modeling, our refinement process is conditioned on the  high-resolution source image $x_s^{HR}$, which serves as a high-fidelity reference to restore details lost during the initial LR editing stage. Specifically, the bridge model learns a conditional velocity field:
\begin{equation}
    v_\theta(X_t, t \mid x_s^{HR}) \approx u_t(X_t \mid \tilde{x}_t^{HR}, x_t^{HR}),
\end{equation}
where the network progressively transforms the noisy state $X_t$ into a high-fidelity result by integrating information from $x_s^{HR}$.

We implement the bridge model using a Diffusion Transformer (DiT)~\cite{peebles2023scalable} backbone. In Fig.~\ref{fig:method}, at each timestep $t$, the model receives the current state $X_t$ along with contextual features extracted from $x_s^{HR}$. The network predicts the velocity term according to the matching objective in Eq.~\eqref{eq:loss}. The final refined image is obtained by integrating the learned probability path from $t=0$ to $t=1$.

Compared to generating high-resolution images from pure noise (Fig.~\ref{fig:motivation}a), the bridge formulation shortens the generative trajectory. Furthermore, explicitly conditioning on the uncorrupted $x_s^{HR}$ ensures the faithful recovery of fine-grained details and high-frequency textures.

\subsection{Prior-Guided Block-wise Sparse Attention}
\label{sec:sparse_attention}
In the context of high-resolution image editing, where the source image is typically integrated as a conditional context via long-sequence concatenation, standard attention mechanisms exhibit a prohibitive quadratic computational complexity $\mathcal{O}(N^2)$ relative to the sequence length $N$. As the token count $N$ scales quadratically with the image resolution (e.g., 2K or 4K), this leads to unsustainable memory footprints and severe latency bottlenecks. To mitigate these challenges, we introduce \textbf{Prior-Guided Block-wise Sparse Attention (PG-BSA)}, a sparsity strategy specifically tailored for high-resolution refinement. Unlike generic sparse attention, PG-BSA leverages the intrinsic semantic alignment between the source and target domains to prune redundant interactions, ensuring both computational efficiency and reconstruction fidelity.

\subsubsection{Attention-Induced Correspondence Prior}

The cornerstone of our method is the utilization of semantic knowledge from the first-stage low-resolution editing. During the initial stage, the model establishes a coarse-grained mapping between the high-resolution source and the edited target. We capture this relationship through the cross-domain attention affinity maps. 

Let $Q_{LR} \in \mathbb{R}^{n_t \times d}$ and $K_{LR} \in \mathbb{R}^{n_s \times d}$ denote the queries from the target domain and keys from the source domain in the first stage, respectively. The attention matrix $A$ is computed as:
\begin{equation}
A = \text{Softmax}\left(\frac{Q_{LR} K_{LR}^\top}{\sqrt{d}}\right) \in \mathbb{R}^{n_t \times n_s},
\end{equation}
where $n_t$ and $n_s$ represent the number of tokens at the native low resolution. To derive a robust spatial prior, we identify the source anchor that yields the maximum response for each target query, effectively establishing a hard correspondence roadmap:
\begin{equation}
\pi(i) = \arg\max_{j} A_{ij}.
\end{equation}
This mapping $\pi(\cdot)$ serves as a dense guidance field. Since the second-stage refinement operates at a significantly higher resolution, we apply a coordinate-based upscaling to  $\hat{\pi}(\cdot)$. Specifically, with an upscaling factor of $S=2$, each low-resolution token $j$ at spatial coordinate $(x, y)$ is mapped to a $2 \times 2$ cluster of high-resolution tokens.  Formally, for any high-resolution token $k$ located at position $(x', y')$, its prior is assigned as $\hat{\pi}(k) = \pi(\lfloor x'/S \rfloor, \lfloor y'/S \rfloor)$. This nearest-neighbor expansion strategy ensures that the semantic guidance remains spatially aligned across scales, providing a robust and piece-wise constant search space for the subsequent prior-guided sparse attention.

\subsubsection{Prior-Guided Sparsification Mechanism}

In the high-resolution refinement stage, the input sequence is structured as a concatenated multi-modal embedding: $[\text{Text} \;|\; \text{Target Image} \;|\; \text{Source Image}]$. To implement the prior-guided sparsification  while preserving global semantic consistency, we decouple the attention mechanism into two distinct computational paths:

\begin{itemize}
    \item \textbf{Intra-domain Self-attention:} To maintain internal structural coherence, we apply localized self-attention independently within the Text ($\mathcal{A}_{txt}$), Target ($\mathcal{A}_{tgt}$), and Source ($\mathcal{A}_{src}$) segments. This domain-isolated processing ensures that each modality focuses on its intrinsic dependencies.
    
    \item \textbf{Prior-guided Cross-domain Attention:} To facilitate information flow, interactions between Target and Source tokens are strictly constrained by the upscaled prior $\hat{\pi}(\cdot)$. For a target query token $q_j$ at position $j$, the attention is computed over a sparse set of source keys:
    \begin{equation}
        \text{Attn}(q_j, K, V) = \sum_{k \in \mathcal{S}_j} \text{Softmax}\left(\frac{q_j k_k^\top}{\sqrt{d}}\right) v_k,
    \end{equation}
    where the receptive field $\mathcal{S}_j$ is defined as a $k \times k$ 
local window centered at the semantically aligned anchor $\hat{\pi}(j)$. This window size compensates for potential spatial 
misalignments in the upscaled prior, with larger windows at higher 
resolutions accounting for increased discretization errors, while 
maintaining $k^2 \ll N$ to ensure computational efficiency.

\end{itemize}

By restricting the interaction to a local neighborhood $\mathcal{S}_j$, we reduce the complexity of cross-domain attention from $\mathcal{O}(N^2)$ to $\mathcal{O}(N \cdot k^2)$. This formulation ensures that each target patch synthesizes high-frequency details by consulting only its most relevant source counterpart, significantly reducing the memory footprint during inference.

\begin{table}[t]
\centering
\caption{Quantitative comparison results. $\uparrow$/$\downarrow$ indicate whether higher/lower values are better. Bold numbers represent the best results.}
\label{tab:main_results}
\small
\renewcommand{\arraystretch}{1.2}
\setlength{\tabcolsep}{6pt}

\begin{tabular}{l r r r r r r}
\toprule
Method
& Haar $\uparrow$
& mPSNR $\uparrow$
& mSSIM $\uparrow$
& mMSE $\downarrow$
& mLPIPS $\downarrow$
& Time (s) $\downarrow$ \\
\midrule

\rowcolor{gray!15}
\multicolumn{7}{l}{\textbf{1K Resolution}} \\
Direct Inference & 0.3973 & 16.8078 & 0.7539 & 0.0549 & 0.4116 & 20.00 \\
DiT4SR~\cite{duan2025dit4sr} & 0.4127 & 14.5031 & 0.7092 & 0.0554 & 0.4818 & 28.47 \\
DiT-SR~\cite{cheng2025effective} & 0.4523 & 16.7152 & 0.7955 & 0.0430 & 0.4403 & 2.44 \\
PiSA-SR~\cite{Sun_2025_CVPR} & 0.4247 & 15.2115 & 0.7481 & 0.0486 & 0.4492 & \textbf{0.30} \\
TSD-SR~\cite{dong2025tsd} & 0.4127 & 15.3864 & 0.7531 & 0.0516 & 0.4628 & 0.42 \\
HiFlow~\cite{bu2025hiflow} & 0.3012 & 11.6964 & 0.6489 & 0.1323 & 0.6012 & 5.12 \\
ScaleEdit~\cite{lee2025lowresolutioneditingneedhighresolution} & 0.4824 & 16.8461 & 0.8206 & 0.0408 & \textbf{0.3784} & 181.20 \\
\textbf{Ours} & \textbf{0.4992} & \textbf{18.6536} & \textbf{0.8686} & \textbf{0.0369} & 0.3829 & 0.84 \\
\addlinespace[0.5em]

\rowcolor{gray!15}
\multicolumn{7}{l}{\textbf{2K Resolution}} \\
Direct Inference & 0.2254 & 13.5262 & 0.7491 & 0.0880 & 0.6700 & 82.14 \\
DiT4SR~\cite{duan2025dit4sr} & 0.3806 & 15.1298 & 0.7468 & 0.0516 & 0.5348 & 113.32 \\
DiT-SR~\cite{cheng2025effective} & 0.4210 & 17.2430 & 0.8265 & 0.0423 & 0.4738 & 14.82 \\
PiSA-SR~\cite{Sun_2025_CVPR} & 0.4046 & 16.2433 & 0.8061 & 0.0446 & 0.4702 & 20.22 \\
TSD-SR~\cite{dong2025tsd} & 0.4046 & 16.2021 & 0.7952 & 0.0467 & 0.4910 & 34.20 \\
HiFlow~\cite{bu2025hiflow} & 0.3144 & 13.9314 & 0.7377 & 0.0759 & 0.5113 & 33.11 \\
ScaleEdit~\cite{lee2025lowresolutioneditingneedhighresolution} & 0.4583 & 16.5510 & 0.6818 & 0.0485 & 0.5197 & 662.50 \\
\textbf{Ours} & \textbf{0.4673} & \textbf{18.6366} & \textbf{0.8712} & \textbf{0.0377} & \textbf{0.3903} & \textbf{4.08} \\
\addlinespace[0.5em]

\rowcolor{gray!15}
\multicolumn{7}{l}{\textbf{4K Resolution}} \\
Direct Inference & 0.2134 & 10.7086 & 0.5411 & 0.1281 & 0.7774 & 695.23 \\
DiT4SR~\cite{duan2025dit4sr} & 0.4002 & 20.2771 & 0.7535 & 0.0133 & 0.3510 & 533.83 \\
DiT-SR~\cite{cheng2025effective} & 0.4267 & 16.8615 & 0.6818 & 0.0415 & 0.5436 & 119.68 \\
PiSA-SR~\cite{Sun_2025_CVPR} & 0.4267 & 21.6200 & 0.7974 & 0.0108 & 0.2687 & 111.42 \\
TSD-SR~\cite{dong2025tsd} & 0.4124 & 22.2766 & 0.8035 & 0.0074 & 0.2399 & 108.72 \\
HiFlow~\cite{bu2025hiflow} & 0.3307 & 14.4400 & 0.6719 & 0.0664 & 0.6038 & 129.30 \\
ScaleEdit~\cite{lee2025lowresolutioneditingneedhighresolution} & 0.4116 & 18.5714 & 0.7638 & 0.0390 & 0.6015 & 3601.20 \\
\textbf{Ours} & \textbf{0.4279} & \textbf{24.1380} & \textbf{0.8347} & \textbf{0.0049} & \textbf{0.2112} & \textbf{61.10} \\
\bottomrule
\end{tabular}
\end{table}

\section{Experiments}
\label{exp}
\subsection{Implementation Details}
\textbf{Training Settings.} We build our proposed method upon Qwen-Image-Edit~\cite{wu2025qwenimagetechnicalreport}, a state-of-the-art image editing model. To realize the sparse attention mechanism, we adopt the VMoBA framework~\cite{wu2025vmobamixtureofblockattentionvideo} and leverage FlashAttention~\cite{dao2022flashattention} to optimize the underlying kernel operations. During training, we employ the Prodigy optimizer~\cite{mishchenko2024prodigyexpeditiouslyadaptiveparameterfree} with a learning rate of $1.0$, a weight decay of $0.01$, enabling a safeguard warmup. Additionally, we apply Low-Rank Adaptation (LoRA)~\cite{hu2022lora} to all linear layers, setting the rank parameter $r=128$ and $\alpha=128$. All experiments are conducted on 8 NVIDIA H800 GPUs (80GB) with a per-GPU batch size of $1$.

\textbf{Dataset.} To construct our training corpus, we curate high-resolution source images from publicly available datasets, including Aesthetic-4k~\cite{zhang2025diffusion} and Aesthetic-Train-V2~\cite{zhang2025ultra}. For images satisfying the resolution criteria, we employ a center-cropping strategy to crop them to the target resolutions. Subsequently, we utilize Gemini 3~\cite{gemini3} to automatically generate detailed editing instructions. The corresponding target images are synthesized via Nano Banana Pro~\cite{gemini_image_pro}. The final dataset consists of $5,000$ image pairs at resolutions of 1K and 2K, supplemented by $1,500$ image pairs at 4K resolution.

\textbf{Evaluation Metrics.} To comprehensively evaluate the performance of our method, we follow the evaluation protocol established by ScaleEdit~\cite{lee2025lowresolutioneditingneedhighresolution}. Specifically, we employ HaarPSI~\cite{Reisenhofer_2018} to directly assess the perceptual similarity between the source and the synthesized images. Furthermore, we incorporate several region-aware metrics to provide a granular analysis of the editing quality. Utilizing a binary mask to bifurcate the image, we compute \textbf{M-PSNR}, \textbf{M-SSIM}, and \textbf{M-MSE} over the \textit{unedited regions} to quantify the fidelity. Conversely, \textbf{M-LPIPS} is restricted to the \textit{edited regions} to evaluate the generative fidelity and perceptual quality of the specific modifications. This dual-region evaluation ensures that our model not only achieves high-quality refinement but also maintains strict consistency in non-target areas.

\subsection{Main Comparison}

We benchmark our method against several state-of-the-art diffusion-based super-resolution (SR) models, including \textbf{DiT-SR} \cite{cheng2025effective}, \textbf{DiT4SR} \cite{duan2025dit4sr}, \textbf{PiSA-SR} \cite{Sun_2025_CVPR}, and \textbf{TSD-SR} \cite{dong2025tsd}.  We further compare our approach with \textbf{ScaleEdit}\cite{lee2025lowresolutioneditingneedhighresolution}, a high-resolution image editing method that employs test-time patch-wise optimization. Additionally, we compare our approach with high-resolution generative models \textbf{HiFlow}~\cite{bu2025hiflow} to underscore the difference between generic upscaling and fidelity-preserving editing. A Direct Inference baseline(using the base model without modifications) is also included.

\begin{figure}[t]
    \centering    \includegraphics[width=1\linewidth]{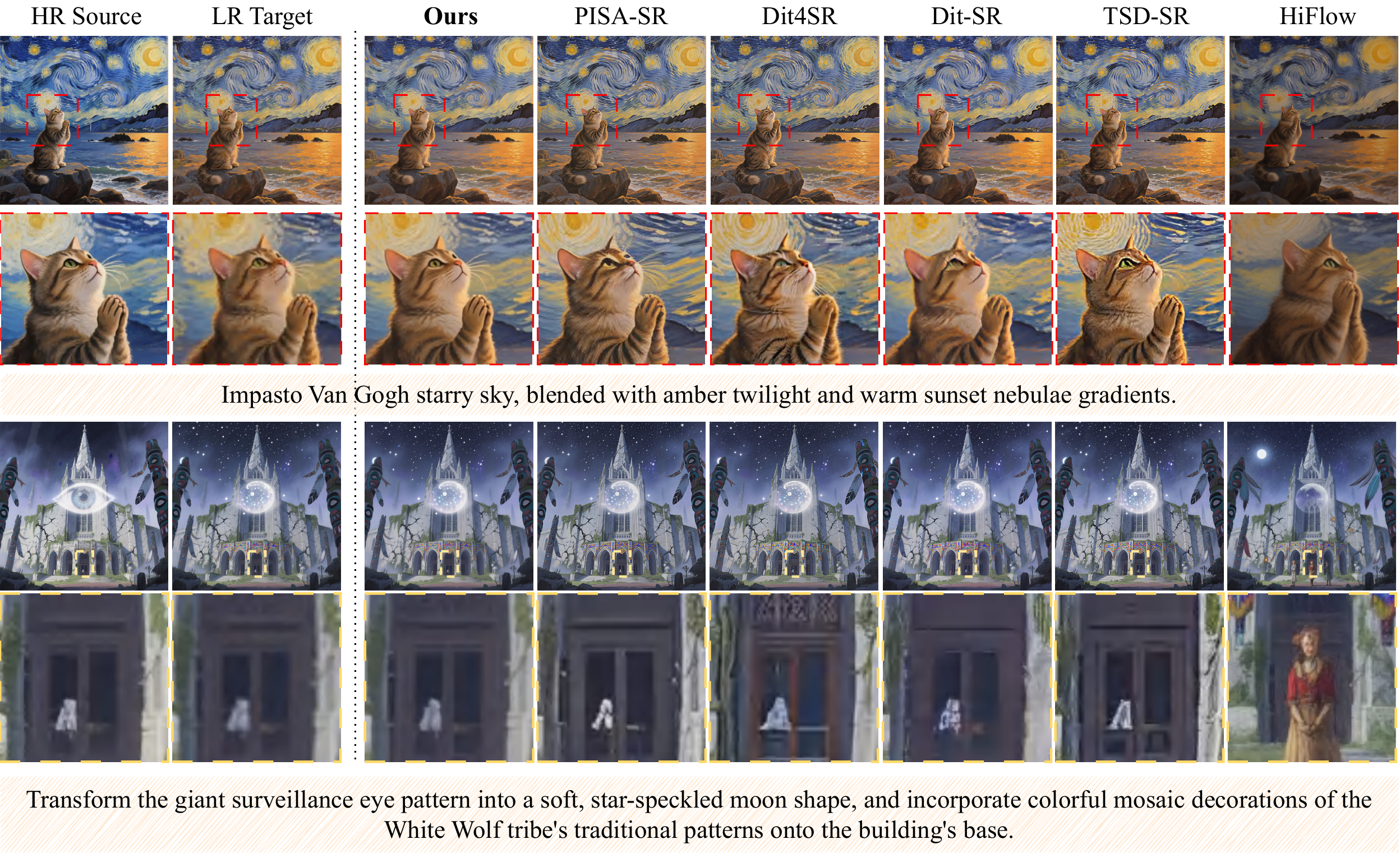}
    \caption{Qualitative comparison at 1K resolution. Our method better preserves fidelity while enhancing visual clarity.}
    \label{fig:1k}
    \vspace{-5mm}
\end{figure}

Tab.~\ref{tab:main_results} presents the quantitative evaluation of our method at 1k, 2k, and 4k resolutions, respectively. As shown, our approach consistently outperforms the baseline models across all resolution settings. Furthermore, we evaluated the inference time of our method, which explicitly accounts for the computational overhead of the prior estimation process. Our method achieves 3.6--8.4$\times$ speedup at 2K compared to 
diffusion-based SR methods, requiring 
only 4.08 seconds. At 4K, while the speedup over SR methods is 
1.8--2.1$\times$, our method achieves 11$\times$ acceleration compared to direct 
high-resolution inference, completing 4K editing in 61 seconds.

\begin{figure}[t]
    \centering
    \includegraphics[width=1\linewidth]{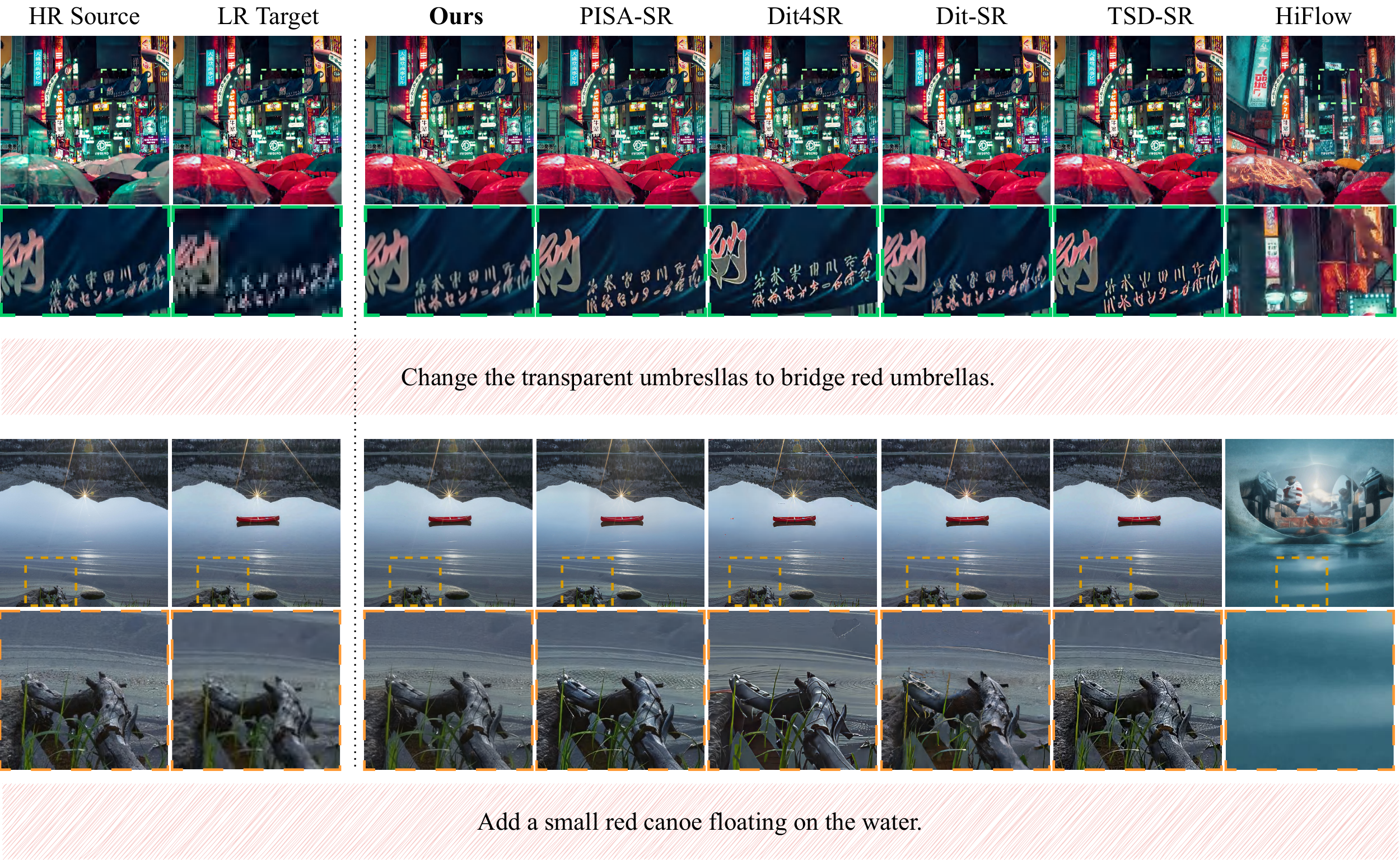}
    \caption{Qualitative comparison at 2K resolution. Our method better preserves fidelity while enhancing visual clarity.}
    \vspace{-5mm}
    \label{fig:2k}
\end{figure}

We further provide a qualitative evaluation of our results. As illustrated in Fig.~\ref{fig:1k}, our method achieves exceptional visual clarity with richer fine-grained details compared to the low-resolution targets, while strictly maintaining the highest structural and semantic consistency with the high-resolution source. Taking the first example in Fig.~\ref{fig:1k}, conventional super-resolution methods suffer from excessive sharpening artifacts, leading to unnatural texture degradation and severely compromised visual fidelity. In the second example, the baseline models frequently hallucinate arbitrary details, causing the generated door to diverge significantly from the original HR source. Furthermore, direct high-resolution visual generation methods tend to alter the overall global semantics of the image. As illustrated in Fig.~\ref{fig:2k}, similar advantages and consistent visual improvements are also observed under the 2k resolution settings.

\begin{table}[b]
\centering
\vspace{-5mm}
\caption{Performance comparison between our PG-BSA and full attention across 1K and 2K settings. $\uparrow$/$\downarrow$ indicate whether higher/lower values are better. Bold numbers represent the best results within each resolution block.}
\label{tab:attention_comparison}
\resizebox{\textwidth}{!}{
\begin{tabular}{llcccccc}
\toprule
Setting & Method & HaarPSI$\uparrow$ & mPSNR$\uparrow$ & mSSIM$\uparrow$ & mMSE$\downarrow$ & mLPIPS$\downarrow$ & Time (s)$\downarrow$ \\ 
\midrule
\multirow{2}{*}{1K} 
& Ours          & 0.499 & 18.653 & \textbf{0.868} & 0.036 & \textbf{0.382} &  0.84  \\
& FullAttention & \textbf{0.539} & \textbf{19.080} & 0.857 & \textbf{0.030} & 0.599 & \textbf{0.73} \\ 
\midrule
\multirow{2}{*}{2K} 
& Ours          & 0.467 & \textbf{18.636} & \textbf{0.871} & \textbf{0.037} & \textbf{0.390} &  \textbf{4.08} \\
& FullAttention &   \textbf{0.505} & 17.918 & 
0.705 & 0.041 & 0.425 & 4.71  \\ 
\bottomrule
\end{tabular}
}
\end{table}

\subsection{Ablation Study}

\begin{figure}[t]
    \centering
    \includegraphics[width=0.95\linewidth]{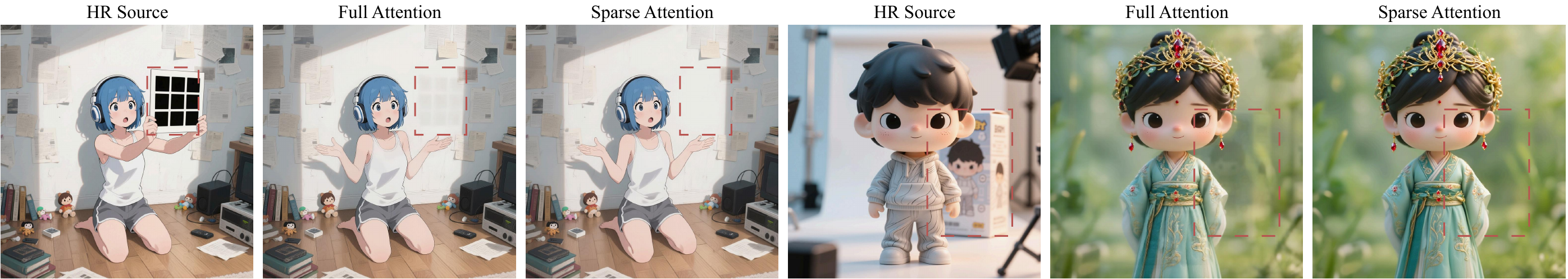}
    \caption{\textbf{Ablation of Attention Sparsity.} Full attention  leads to noticeable source-induced artifacts, whereas our sparse attention maintains superior visual fidelity and alignment.}
    \vspace{-5mm}
    \label{fig:attention_visual}
\end{figure}

\textbf{Sparse Attention vs. Full Attention.} We further investigate the impact of different attention mechanisms at 1K and 2K resolutions to validate our design. As shown in Tab.~\ref{tab:attention_comparison}, Full Attention achieves slightly higher quantitative scores in pixel-level reconstruction metrics (e.g., PSNR, HaarPSI). However, this quantitative superiority can be deceptive in the context of local editing. Since full attention computes global interactions without spatial constraints, the model tends to indiscriminately over-attend to semantically irrelevant tokens from the high-resolution source image. This contextual bleeding introduces noticeable ghosting and structural artifacts in the newly synthesized areas, as illustrated in Fig.~\ref{fig:attention_visual}.

Moreover, the quadratic computational complexity of global attention imposes severe memory footprints and latency bottlenecks, rendering it computationally prohibitive for high-resolution inference. In contrast, our proposed sparse attention explicitly constrains cross-image interactions using semantic registration priors. By routing queries strictly to their semantically aligned anchors, it effectively prunes noisy global contexts. This design not only curtails the computational complexity to a linear scale—yielding the substantial speedups discussed earlier—but also strictly preserves local detail fidelity without unwanted interference from unedited regions.

\begin{figure}[h]
    \centering
    \includegraphics[width=0.95\linewidth]{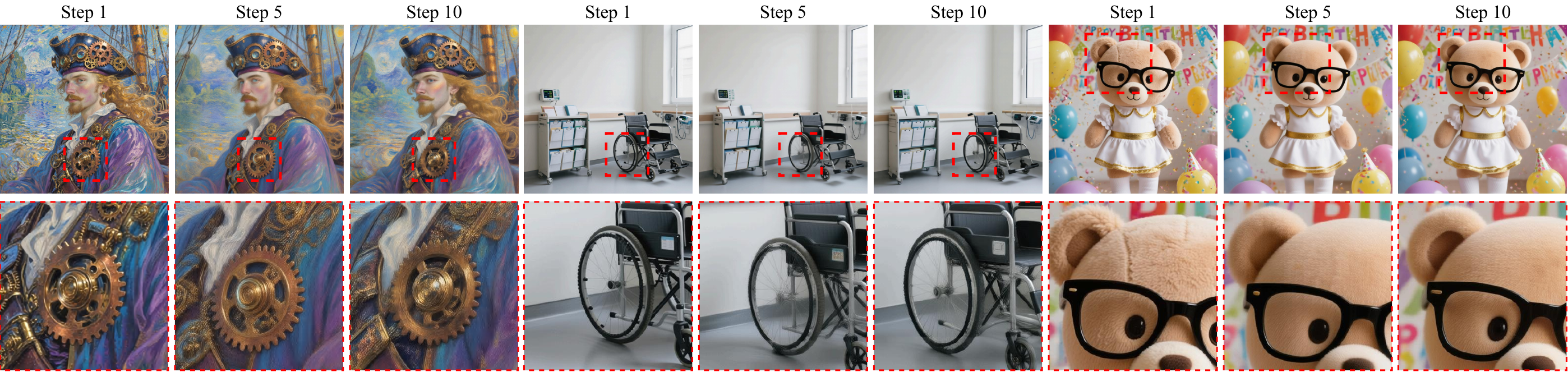}
    \caption{\textbf{Impact of sampling steps.} Perceptual quality and high-frequency details are maximized at $T=1$. Increasing the number of inference steps does not yield significant visual gains, further validating the efficiency of our refinement framework.}
    \label{fig:ablation}
\end{figure}

\textbf{Impact of Inference Steps.} We evaluate the model performance across varying numbers of inference steps ($N$), as illustrated in Tab.~\ref{tab:ablation_steps}. Interestingly, empirical results indicate that single-step inference ($N=1$) achieves the optimal trade-off between computational efficiency and perceptual quality. Unlike multi-step sampling, which may accumulate quantization errors at extreme resolutions, single-step inference yields the highest fidelity to the source's high-frequency structures while maintaining minimal computational cost (e.g., a mere 0.84s for 1K images). Furthermore, as illustrated in Fig.~\ref{fig:ablation}, the setting $N=1$ enables the model to exhibit the richest fine-grained details.

\begin{table}[h]
\vspace{-3mm}
\centering
\caption{Quantitative evaluation of varying inference steps. $\uparrow$/$\downarrow$ indicate whether higher/lower values are better. Bold numbers represent the best results.}
\label{tab:ablation_steps}
\resizebox{0.95\textwidth}{!}{
\begin{tabular}{lcccccc}
\toprule
Method & HaarPSI$\uparrow$ & mPSNR$\uparrow$ & mSSIM$\uparrow$ & mMSE$\downarrow$ & mLPIPS$\downarrow$ &  Time(s)$\downarrow$ \\ 
\midrule
5 steps  & 0.4131 & 15.7269 & 0.7763 & 0.0474 & 0.4739  & 4.03 \\
10 steps & 0.4020 & 15.4190 & 0.7636 & 0.0494 & 0.4662 &  8.10 \\
\textbf{Ours (1 step)} & \textbf{0.4991} & \textbf{18.6535} & \textbf{0.8685} & \textbf{0.0368} & \textbf{0.3829} & \textbf{0.84} \\
\bottomrule
\end{tabular}
}
\end{table}

\vspace{-5mm}
\section{Conclusion}
In this paper, we introduced \textbf{EditBridge}, a novel prior-guided diffusion bridge framework tailored for ultra-high-resolution image editing. Our approach formulates high-resolution refinement as a continuous, data-to-data translation process. By leveraging the original high-resolution source image as conditional guidance, EditBridge overcomes the critical limitations of conventional two-stage super-resolution pipelines, namely information divergence and texture hallucination, thereby ensuring faithful detail preservation. Furthermore, to address the prohibitive computational costs associated with high-resolution processing, we proposed a task-specific, prior-guided sparse attention mechanism. By dynamically routing information based on spatial semantic alignment extracted from the low-resolution editing stage, our model selectively focuses only on relevant high-resolution anchors, effectively pruning redundant global interactions.  Overall, our framework establishes a highly efficient and scalable paradigm for high-fidelity image editing.

\newpage

{\small
\bibliographystyle{unsrtnat}
\bibliography{main}
}

\newpage
\appendix
\appendixpage
\startcontents[sections]
\printcontents[sections]{l}{1}{\setcounter{tocdepth}{2}}
\newpage
\section{User study}
\label{sec:user_study}

\begin{figure}[h]
    \centering
    \includegraphics[width=1\linewidth]{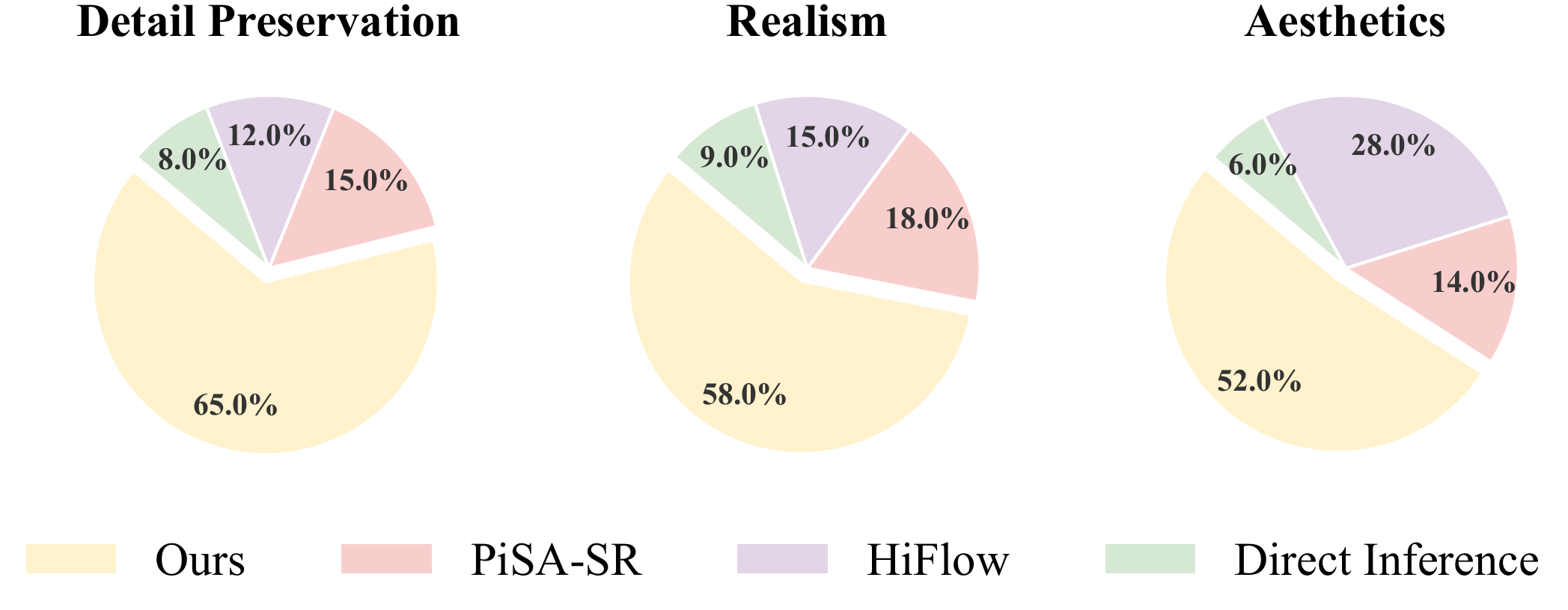}
    \caption{Results of the user study across three dimensions: \textit{Detail Preservation}, \textit{Realism}, and \textit{Aesthetics}. Our method consistently outperforms all baselines in human preference. }
    \label{fig:user_study}
\end{figure}

To evaluate subjective visual quality beyond quantitative metrics, we conducted a blind Four-Alternative Forced Choice user study. Thirty participants evaluated 50 randomly sampled image pairs. We compared EditBridge against direct inference and PiSA-SR, HiFlow representing SR method, high-resolution visual generation method, respectively. Participants were provided with both full images and 100\% cropped patches and voted based on three criteria: \textit{Detail Preservation} (fidelity to unedited high-frequency source textures), \textit{Realism} (naturalness of micro-structures), and \textit{Aesthetics} (overall composition and structural integrity). 

As shown in Fig.~\ref{fig:user_study}, EditBridge consistently achieves the highest preference rates across  all evaluated dimensions. Specifically, our method significantly outperforms PiSA-SR and HiFlow in Detail Preservation and Realism. This confirms that our prior-guided sparse attention effectively prevents the texture hallucination typical of SR pipelines and the information divergence common in pure generative models. Furthermore, while HiFlow exhibits strong performance in the \textit{Aesthetics} dimension—owing to the remarkable aesthetic priors inherent in large generative models—EditBridge still secures the highest overall aesthetic preference. This demonstrates that our approach successfully marries superior visual aesthetics with strict structural fidelity.

\section{Training Configuration}
\label{Training_config}
We provide the detailed training configuration below.

\begin{tcolorbox}[
    title=Training Configuration,
    colframe=black!10,
    colback=white,
    coltitle=black,
    fonttitle=\bfseries\sffamily,
    breakable,
    left=5mm
]
\begin{lstlisting}[
    basicstyle=\small\ttfamily,
    breaklines=true,
    columns=fullflexible,
    xleftmargin=0em,
    literate={:}{{\textcolor{red}{:}}}1
]
lora_config:
  r: 128
  lora_alpha: 128
  init_lora_weights: "gaussian"
  target_modules: "all-linear"

optimizer:
  type: "Prodigy"
  params:
    lr: 1
    use_bias_correction: true
    safeguard_warmup: true
    weight_decay: 0.01
\end{lstlisting}
\end{tcolorbox}

\section{Pseudo Code}
We present our training procedure in Algorithm \ref{alg:training} and the inference sampling process in Algorithm \ref{alg:denoising_sparse}.

For training, we learn a velocity field that bridges the ground truth distribution and the conditional distribution. Specifically, we define the ground truth latent $z_{gt}$ as the flow source $z_0$ and the low-resolution target $z_{lr\_tgt}$ as the flow target $z_1$. We sample a timestep $t$ and construct a fused latent $z_t$ through the Bridge forward process to train the model to predict the corresponding vector field. To preserve high-frequency details, we concatenate the high-resolution source $z_{hr}$ along the sequence dimension. We employ a sparse attention mechanism where the attention mask $\Omega$ is pre-computed based on the correspondence between the low-resolution versions of the source and target, which regularizes the model during optimization of the weighted MSE loss.

For inference, we perform progressive denoising. Starting from the initial high-resolution source latent $z_1$, we iteratively solve the stochastic differential equation (SDE) toward the target state $z_0$. In each denoising step, we downsample the current latent $z_t$ to re-estimate the sparse attention mask $\Omega$, ensuring that the Transformer operates under accurate structural constraints relative to the fixed high-resolution reference. We incorporate a Brownian Bridge update rule, where a predicted velocity field and a scaled noise term are combined to update the latent state. This sampling strategy, stabilized by a noise rescaling factor, enables high-fidelity reconstruction of the edited image while adhering to the input text and reference source.

\SetKwInput{Input}{Input}
\SetKwInput{Output}{Output}
\SetKw{Continue}{continue}

\begin{algorithm}[t]
    \caption{Training Procedure of Resolution-Bridge Model with Sparse Attention}
    \label{alg:training}
    
    \Input{High-res Source $I_{hr}$, Low-res Target $I_{lr\_tgt}$, Ground Truth $I_{gt}$, Text Prompt $y$}
    \Output{Optimized Loss $\mathcal{L}$}

    \BlankLine

    $I_{lr\_src} \leftarrow \text{Downsample}(I_{hr})$ 

    \BlankLine

    $z_{hr}, z_{gt}, z_{lr\_tgt}, z_{lr\_src} \leftarrow \text{VAE}(I_{hr}, I_{gt}, I_{lr\_tgt}, I_{lr\_src})$\;
    $c \leftarrow \text{TextEncoder}(y, z_{lr\_tgt})$\;
    
    \BlankLine

    $z_0 \leftarrow z_{gt}$\; 
    $z_1 \leftarrow z_{lr\_tgt}$\; 
    
    \BlankLine

    Sample time $t \sim \mathcal{U}(0, 1)$ and noise $\epsilon \sim \mathcal{N}(0, \mathbf{I})$\;
    $z_t \leftarrow (1-t)z_0 + t z_1 + \sigma_t \epsilon$ 
    
    \BlankLine

    $x_{in} \leftarrow \text{Concat}([z_t, z_{hr}], \text{dim}=\text{sequence})$\; 
    $\Omega \leftarrow \text{ExtractSparseIndices}(z_{lr\_tgt}, z_{lr\_src})$\; 
    
    \BlankLine

    $v_\theta \leftarrow \text{Transformer}_\theta(x_{in}, t, c \ ; \Omega)$\;
    $\mathcal{L}_{mse} \leftarrow \left\| v_\theta - \frac{z_t - z_0}{t} \right\|^2$\;
    
    \BlankLine

    \If{stabilized}{
        $w_t \leftarrow \frac{t \|z_1 - z_0\|^2}{t \|z_1 - z_0\|^2 + (1-t)}$\;
        $\mathcal{L} \leftarrow w_t \cdot \mathcal{L}_{mse}$\;
    }
    \Else{
        $\mathcal{L} \leftarrow \mathcal{L}_{mse}$\;
    }
    
    \Return $\mathcal{L}$
\end{algorithm}

\begin{algorithm}[H]
    \caption{Inference Procedure of  Model with Sparse Attention}
    \label{alg:denoising_sparse}
    
    \KwIn{Initial latent $z_1$, High-res reference $z_{hr}$, Low-res source $z_{lr\_src}$, Text embedding $c$, Timesteps $\mathcal{T} = \{t_1, \dots, t_N\}$, Noise scale $\sigma$, Rescale flag $rescale\_noise$}
    \KwOut{Denoised final latent $z_0$}

    \BlankLine
    
    $z_t \leftarrow z_1$\;
    
    \BlankLine
    
    \For{$i = 1$ \KwTo $N$}{
        $t \leftarrow \mathcal{T}[i]$\;
        $t_{next} \leftarrow \mathcal{T}[i+1]$ (or $0$ if $i=N$)\;
        $\Delta t \leftarrow t_{next} - t$\;
        
        \BlankLine
        
        $z_{lr\_tgt} \leftarrow \text{Downsample}(z_t)$\;
        $\Omega \leftarrow \text{ExtractSparseIndices}(z_{lr\_tgt}, z_{lr\_src})$\;
        
        \BlankLine
        
        $x_{in} \leftarrow \text{Concat}([z_t, z_{hr}], \text{dim}=\text{sequence})$\;
        $v_\theta \leftarrow \text{Transformer}_\theta(x_{in}, t, c \ ; \Omega)$\;
        
        \BlankLine
        
        $\eta \leftarrow \sqrt{-\Delta t \cdot \frac{t_{next}}{t}}$\;
        $\epsilon \sim \mathcal{N}(0, \mathbf{I})$\;
        
        \BlankLine
        
        \eIf{rescale\_noise}{
            $\gamma \leftarrow \text{Clip}(|v_\theta|, 0, 1)$\;
        }{
            $\gamma \leftarrow 1$\;
        }
        
        \BlankLine
        
        $z_t \leftarrow z_t + v_\theta \cdot \Delta t + \sigma \cdot \eta \cdot \gamma \cdot \epsilon$\;
    }
    
    \BlankLine
    
    \Return $z_t$
\end{algorithm}

\section{Prompts}

We present detailed prompts used in our experiments below.  The  prompts correspond to the cases shown in the main paper, ordered by their appearance.

\begin{tcolorbox}[
title=Prompts for Editing,
colframe=black!20,
colback=white,
coltitle=black,
fonttitle=\bfseries,
breakable
]

\footnotesize

\textbf{System Prompt Input to Gemini}

\medskip

You are an expert image editing instruction writer.

\medskip
IMPORTANT
\begin{itemize}
\item You must output STRICT JSON.
\item Do NOT output any explanation outside JSON.
\item Do NOT use markdown.
\item Do NOT include text before or after JSON.
\item If you violate this format, the output is invalid.
\end{itemize}

You are given one image.

Your task: generate \textbf{ONE high-quality image editing prompt} based on this image.  
The prompt should describe \textbf{plausible, clear, and specific edits}.  
Keep it in \textbf{one sentence}. The instruction should be in \textbf{English and concise}.

Example:  
``Change the background to a sunny day with blue sky and white clouds.''

\medskip
Final output format

\begin{center}
\ttfamily
\{
\quad "edit\_prompt": "your single-sentence editing prompt"
\}
\end{center}

\medskip
\textbf{Editing Instructions}

\begin{enumerate}
\item Transform the scene into an elementary school hallway with all doors wide open. Show the subject in a confused state, and add question marks above their head.

\item Transform the fox's monochrome fur into a red gradient, transitioning from deep amber to light coral. Position its front paws to be pinching a Nepalese dumpling, featuring a stylized fox-face motif rendered in white ink on the dumpling's skin.

\item Repaint the starry sky scene in a thick impasto oil painting style. Retain Van Gogh's \emph{Starry Night} brushwork while blending the sky with amber dusk hues, forming a warm sunset gradient within the swirling nebulae.

\item Soften the giant surveillance eye motif into a starlit moon shape. Add colorful mosaic decorations with traditional White Wolf tribe patterns to the building base, and lower the overall color saturation to create a melancholic Eastern classical atmosphere.

\item Change the transparent umbrellas to bright red umbrellas.

\item Add a small red canoe floating on the water.

\item Remove the square card the character is holding.

\item Transform the main character from a Chibi (Q-version) style to a classical Eastern aesthetic. Add vine-like gold ornaments and red gemstones to the hair and accessories, replace the clothing with a cyan Hanfu featuring intricate embroidery, and adjust the environment to a natural green-toned setting with shallow depth of field.

\item Create an impressionistic portrait of a Steampunk Pirate Captain combining the blue-violet palette of Monet's \emph{Water Lilies} with Van Gogh’s swirling brushstrokes. Metallic gears should shimmer with cadmium yellow impasto while steampunk structures with starry gradients appear within the background foliage.

\item Seamlessly integrate a second image of a bedside monitor into the ward scene with automatically matched lighting and perspective.

\item Dress the bear in a classic maid-inspired outfit featuring a white apron with gold piping and white stockings. Decorate the environment with balloons and birthday banners to create a whimsical fairytale party atmosphere.
\end{enumerate}

\end{tcolorbox}

\section{Limitation and Future Work}
\label{sec:future_work}
\paragraph{Limitations} 
Despite the superior editing performance demonstrated by our framework, several constraints remain to be addressed. First, our method currently relies on a pre-defined indices prior, which must be extracted or synthesized before the translation process. This dependency introduces a bottleneck in fully automated pipelines, as the quality of the final output is inherently bounded by the precision of these initial guidance signals. Second, the iterative nature of the Diffusion Bridge sampling process, while robust, still incurs a non-negligible computational overhead, particularly when dealing with high-resolution imagery or complex multi-step transitions.

\paragraph{Future Work: Automated Prior Estimation} 
To enhance the autonomy of our approach, a promising future direction is to integrate a learnable prior estimation module. By leveraging self-supervised representation learning, we aim to enable the model to infer the necessary structural or semantic indices directly from the source image, thereby eliminating the need for manual or external prior acquisition. This would significantly broaden the applicability of Diffusion Bridge models in "in-the-wild" image editing scenarios.

\paragraph{Future Work: Post-hoc Refinement and Localized Repair} 
Beyond global image-to-image translation, we envision the Diffusion Bridge as a powerful tool for post-hoc refinement and interactive content repair. Specifically, if certain regions of a generated image suffer from "mode collapse" or textural artifacts, the Diffusion Bridge can be re-initialized within a masked area to perform localized resampling. By bridging the intermediate latent states of a suboptimal generation back to the clean data manifold, the model can ``correct" specific flaws while preserving the global coherence of the scene. 

\paragraph{Generalizing the Bridge Mechanism} 
Furthermore, we plan to investigate the potential of the Diffusion Bridge in broader domains, such as video-to-video editing and 3D asset stylization. Extending the bridge trajectory to maintain temporal and spatial consistency across frames or viewpoints remains a challenging yet rewarding frontier. We believe that the mathematical elegance of the Diffusion Bridge offers a versatile foundation for a wide range of controllable generative tasks.

\section{More Cases}

\begin{figure}[ht]
    \centering
    \includegraphics[width=1\linewidth]{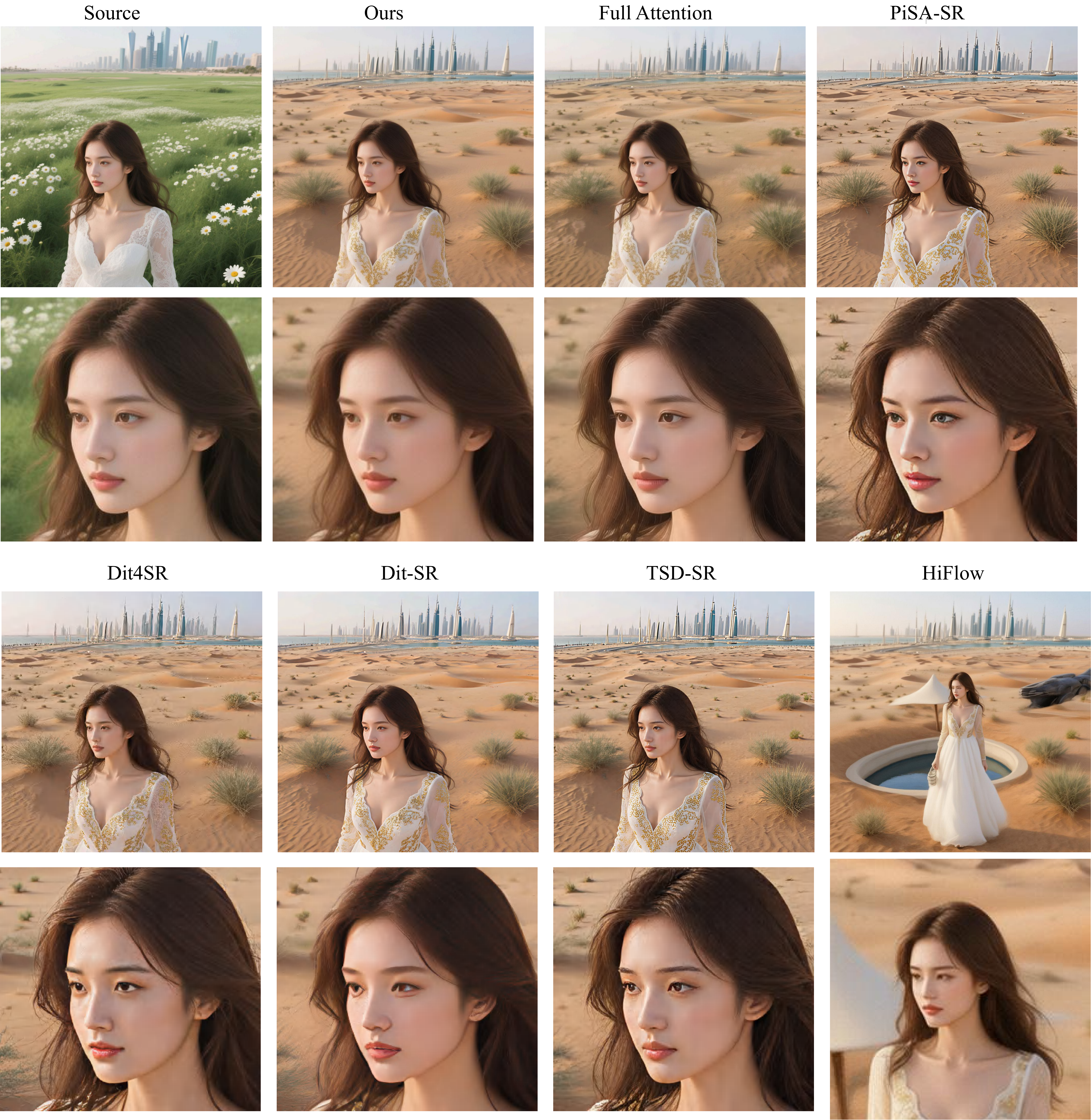}
    \caption{Visual results at 1K resolution.}
    \label{fig:sm2}
\end{figure}

\begin{figure}
    \centering
    \includegraphics[width=1\linewidth]{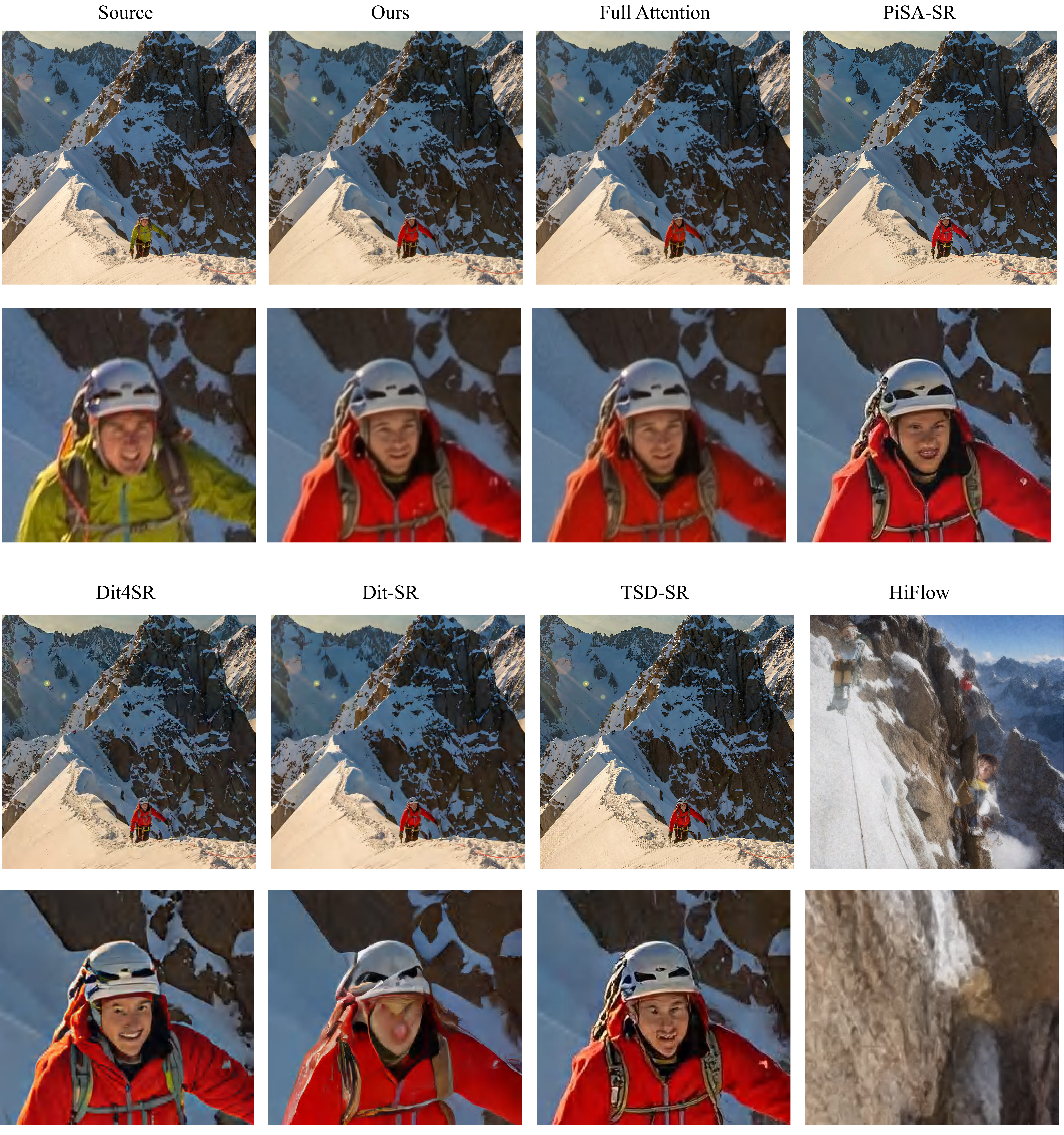}
    \caption{Visual results at 2K resolution.}
    \label{fig:2k_sup}
\end{figure}

We provide an extensive collection of additional samples to further validate the efficacy of our proposed method. As illustrated in Fig.~\ref{fig:sm2} and Fig.~\ref{fig:2k_sup}, our results achieve fidelity to the source images while maintaining the utmost perceptual naturalness in the edited regions. 

Our Diffusion Bridge framework demonstrates a superior ability to preserve the original structural integrity and identity of the subjects. Meanwhile, the synthesized content integrates seamlessly with the existing background without introducing common generative artifacts such as blurriness or texture repetition. These extensive cases substantiate that our approach strikes an optimal balance between strictly following complex textual instructions and faithfully retaining the essential characteristics of the source content.

\end{document}